\documentclass[a4paper,conference]{IEEEtran}
\usepackage{cite}
\usepackage[pdftex]{graphicx}
\usepackage{amsmath}
\usepackage{epsfig}
\usepackage{graphicx}
\usepackage{amssymb}
\usepackage[table,x11names]{xcolor}

\newsavebox\CBox
\def\textBF#1{\sbox\CBox{#1}\resizebox{\wd\CBox}{\ht\CBox}{\textbf{#1}}}

\newcommand{\eg}{{\emph{e.g.}}}

\begin{document}

%
\title{PyNet-V2 Mobile: Efficient On-Device Photo Processing With Neural Networks}


\author{\IEEEauthorblockN{Andrey Ignatov\IEEEauthorrefmark{1}\IEEEauthorrefmark{2},
Grigory Malivenko\IEEEauthorrefmark{2},
Radu Timofte\IEEEauthorrefmark{1}\IEEEauthorrefmark{2},
Yu Tseng\IEEEauthorrefmark{3},
Yu-Syuan Xu\IEEEauthorrefmark{3},
Po-Hsiang Yu\IEEEauthorrefmark{3},\\
Cheng-Ming Chiang\IEEEauthorrefmark{3},
Hsien-Kai Kuo\IEEEauthorrefmark{3},
Min-Hung Chen\IEEEauthorrefmark{3},
Chia-Ming Cheng\IEEEauthorrefmark{3} and
Luc Van Gool\IEEEauthorrefmark{1}\IEEEauthorrefmark{2}}
\IEEEauthorblockA{\IEEEauthorrefmark{1}Computer Vision Laboratory, ETH Zurich, Switzerland}
\IEEEauthorblockA{\IEEEauthorrefmark{2}AI Witchlabs Ltd., Zollikerberg, Switzerland}
\IEEEauthorblockA{\IEEEauthorrefmark{3}MediaTek Inc., Hsinchu, Taiwan}
}


\maketitle

\begin{abstract}
The increased importance of mobile photography created a need for fast and performant RAW image processing pipelines capable of producing good visual results in spite of the mobile camera sensor limitations. While deep learning-based approaches can efficiently solve this problem, their computational requirements usually remain too large for high-resolution on-device image processing. To address this limitation, we propose a novel PyNET-V2 Mobile CNN architecture designed specifically for edge devices, being able to process RAW 12MP photos directly on mobile phones under 1.5 second and producing high perceptual photo quality. To train and to evaluate the performance of the proposed solution, we use the real-world Fujifilm UltraISP dataset consisting on thousands of RAW-RGB image pairs captured with a professional medium-format 102MP Fujifilm camera and a popular Sony mobile camera sensor. The results demonstrate that the PyNET-V2 Mobile model can substantially surpass the quality of tradition ISP pipelines, while outperforming the previously introduced neural network-based solutions designed for fast image processing. Furthermore, we show that the proposed architecture is also compatible with the latest mobile AI accelerators such as NPUs or APUs that can be used to further reduce the latency of the model to as little as 0.5 second. The dataset, code and pre-trained models used in this paper are available on the project website: https://github.com/gmalivenko/PyNET-v2
\end{abstract}


%
\IEEEpeerreviewmaketitle

\section{Introduction}

\begin{figure}[t!]
\centering
\resizebox{0.75\linewidth}{!}{\includegraphics[width=1.0\linewidth]{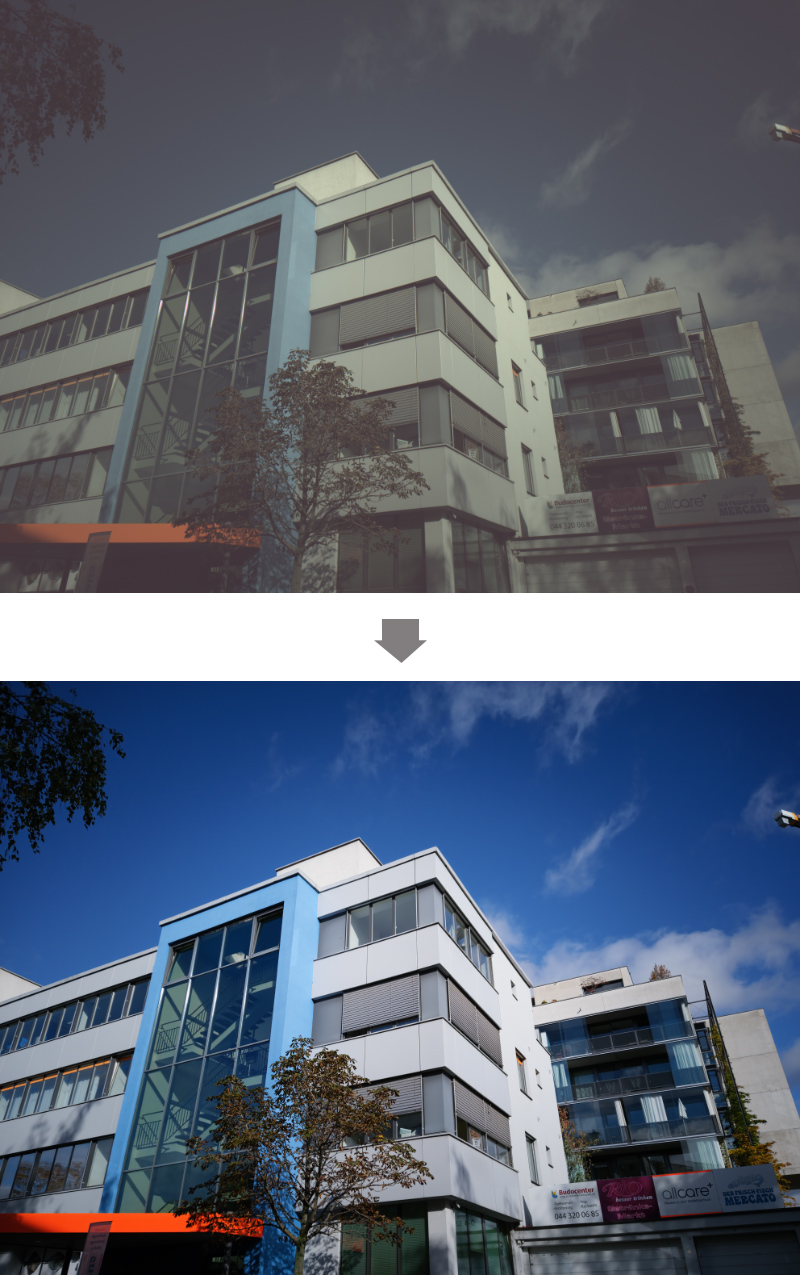}}
\vspace{0.1cm}
\caption{The original visualized RAW photo and the image reconstructed with the proposed PyNet-V2 Mobile model.}
\label{fig:Examples1}
\vspace{-0.3cm}
\end{figure}

\begin{figure*}[t!]
\centering
\setlength{\tabcolsep}{1pt}
\resizebox{\linewidth}{!}
{
\begin{tabular}{ccc}
\scriptsize{Visualized RAW Image}\normalsize & \scriptsize{MediaTek Dimensity 820 ISP Photo}\normalsize & \scriptsize{Fujifilm GFX 100 Photo}\normalsize\\
    \includegraphics[width=0.33\linewidth]{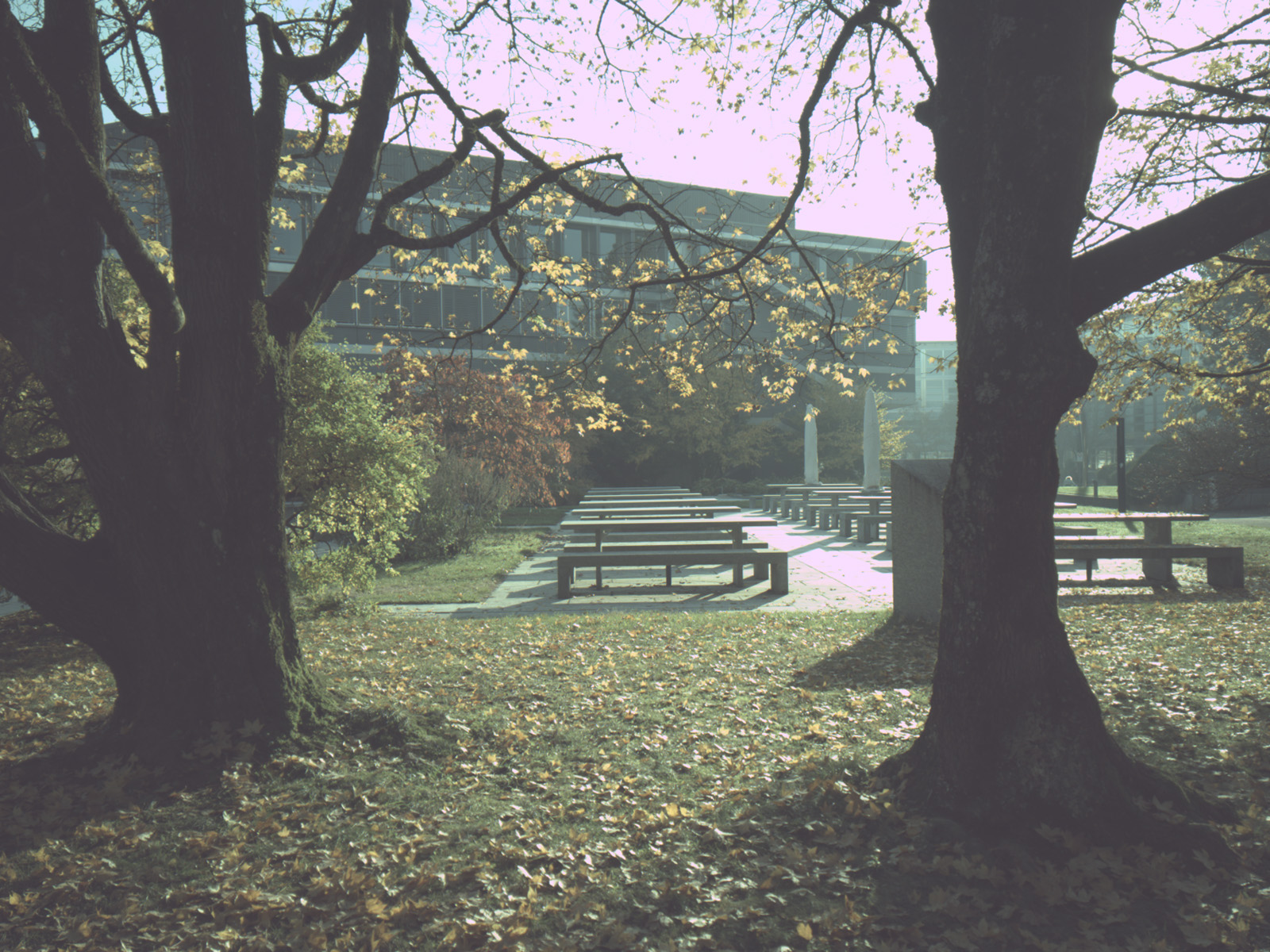}&
    \includegraphics[width=0.33\linewidth]{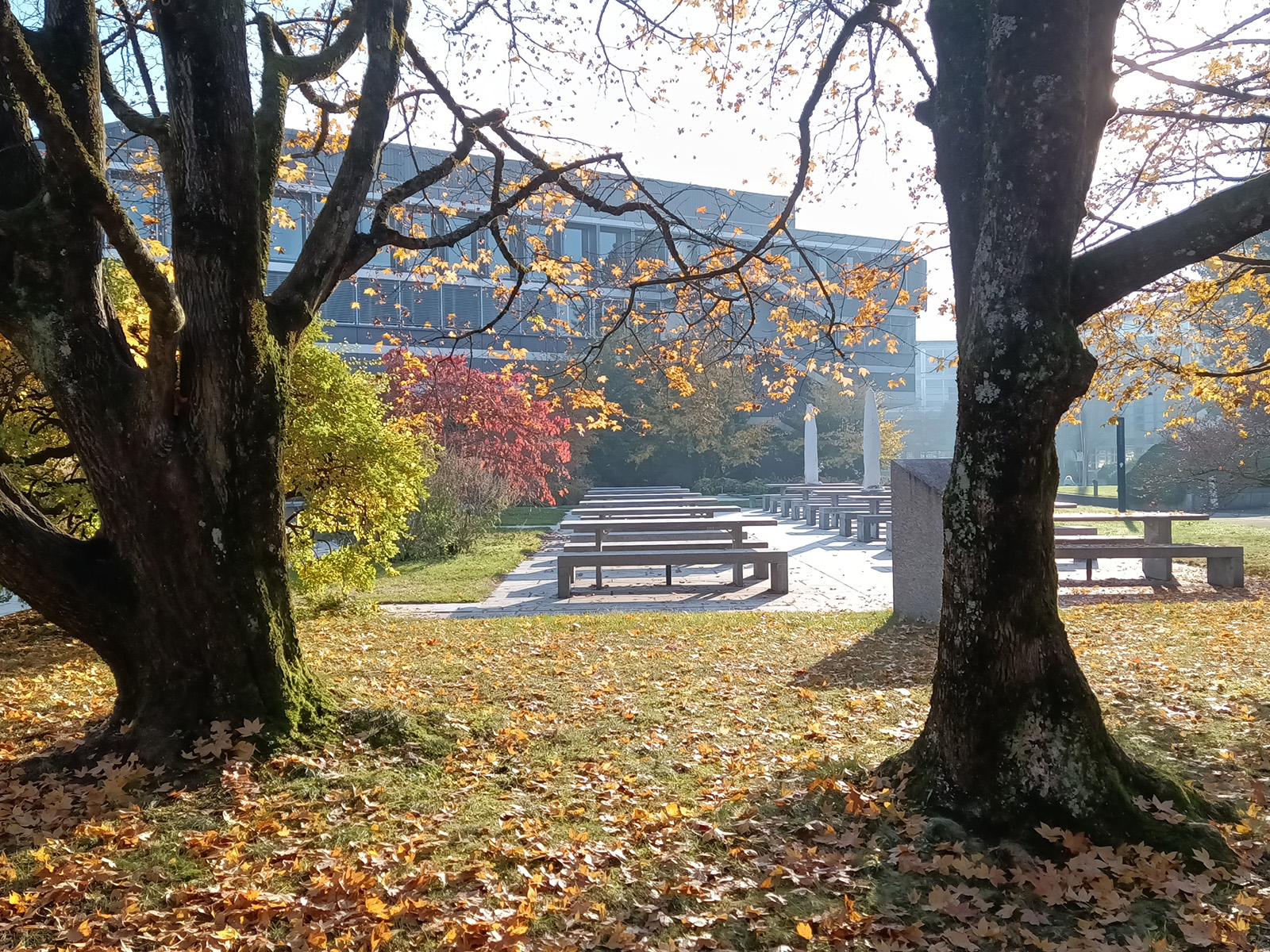}&
    \includegraphics[width=0.33\linewidth]{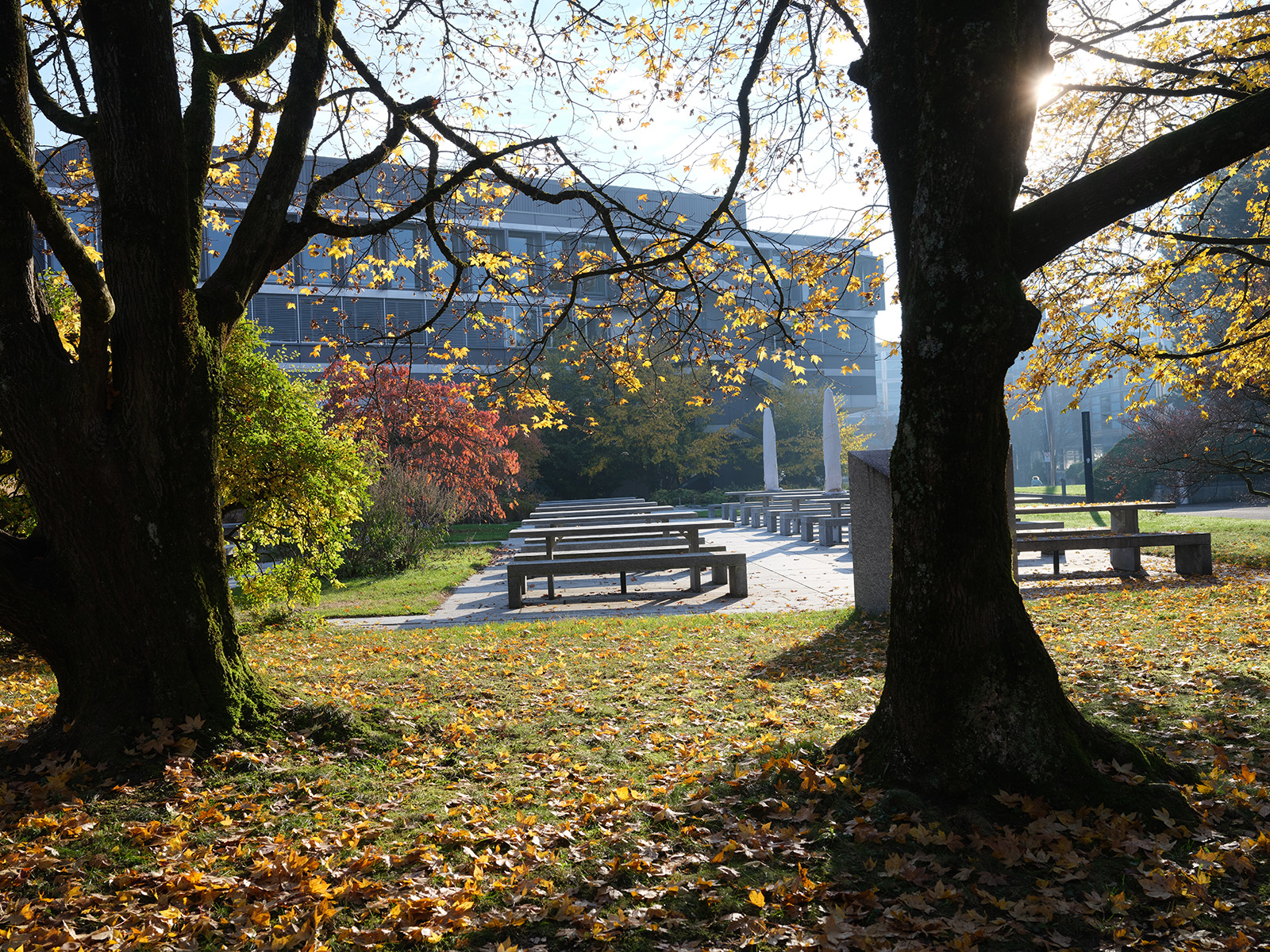} \\
    \includegraphics[width=0.33\linewidth]{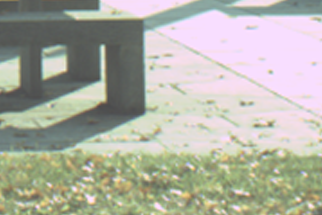}&
    \includegraphics[width=0.33\linewidth]{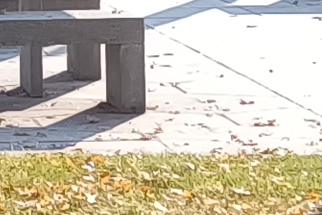}&
    \includegraphics[width=0.33\linewidth]{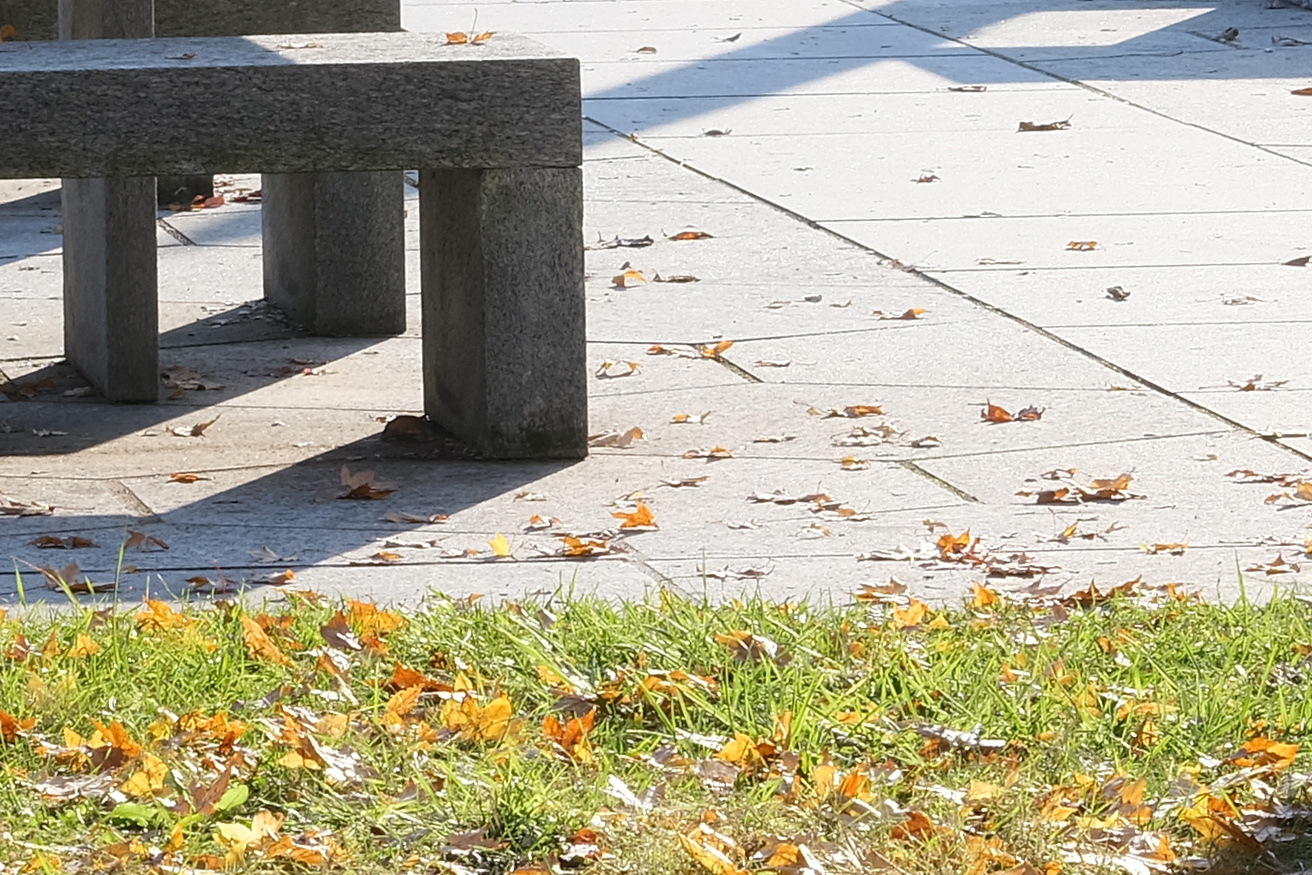} \\
\end{tabular}
}
\vspace{-0.0cm}
\caption{Example set of full-resolution images (top) and crops (bottom) from the collected Fujifilm UltraISP dataset. From left to right: original RAW visualized image, RGB image obtained with MediaTek's built-in ISP system, and Fujifilm GFX100 target photo.}
\vspace{-0.2cm}
\label{fig:dataset}
\end{figure*}

During the past years, mobile devices became a major source of photos taken by regular users, replacing compact point-and-shoot cameras and entry-levels DSLRs. Thus, the demand for high quality smartphone photos has also increased significantly. Lots of efforts are now being devoted to designing powerful image signal processing (ISP) pipelines capable of dealing with hardware limitations of small mobile camera sensors. As the conventional hand-crafted approaches are no longer able to provide a significant boost of image quality, more and more attention is now being paid to deep learning-based computational photography allowing to push the visual results of the processed images to a new level.

The problem of deep learning-based photo enhancement has been addressed in many works, though at the beginning they were dealing only with some narrow image enhancement aspects such as image denoising~\cite{gu2019brief,tai2017memnet,zhang2018ffdnet,zhang2017beyond,abdelhamed2020ntire,abdelhamed2019ntire,ignatov2021fast}, color~\cite{salih2012tone,ma2014high,yan2016automatic,lee2016automatic} and luminance~\cite{yuan2012automatic,fu2016fusion,cai2018learning} adjustments, or resolution improvement~\cite{dong2015image,kim2016accurate,lim2017enhanced,timofte2017ntire,timofte2018ntire,ignatov2018pirm,cai2019ntire,lugmayr2020ntire,zhang2020ntire,ignatov2021real}. However, when dealing with real photos, one generally wants to have a one-in-all approach allowing to improve different quality aspects simultaneously. More importantly, this approach should also be able to take into account the particularities of real data, and not of synthetically generated degraded or enhanced images. This problem was first considered in~\cite{ignatov2018pirm,ignatov2019ntire}, where the authors proposed to learn an end-to-end deep learning based solution for mapping low-quality RGB smartphone photos to images captured with a professional DSLR camera, and presented the DPED dataset consisting of such image pairs for this task. The subsequent works~\cite{vu2018fast,lugmayr2019unsupervised,de2018fast,hui2018perception,huang2018range,liu2018deep,ignatov2018pirm,ignatov2019ntire} have significantly improved the results on this problem and dataset, though one key limitation related to the task definition itself remained: when working with images processed by smartphones' built-in image signal processing pipelines, one has access only to altered pixel data obtained after all image processing steps done by an ISP. In particular, the dynamic range of the resulting photos is lowered from 10 / 12 to 8 bits, many details are washed out during image denoising step, while the texture is altered with various sharpening filters. Thus, one would ideally want to train the model on the original RAW photos that contain much more information that can be potentially used for image enhancement. This was done in~\cite{ignatov2020replacing}, where the dataset and the PyNET model were proposed. The authors claimed that they were able to achieve the results similar to the ones of the ISP system of the Huawei P20 smartphone, while even better visual quality was obtained in subsequent works~\cite{ignatov2020aim,dai2020awnet,silva2020deep,kim2020pynet,ignatov2019aim}. Despite the great results, all these solutions suffered from one major limitation: the proposed models were too heavy for deployment on mobile devices, and required up to tens of seconds even when processing high-resolution photos on high-end desktop GPUs. This problem was partially addressed in the MAI challenge~\cite{ignatov2021learned} targeting efficient deep learning-based ISP systems, though the target image quality and resolution requirements used in this competition were still far from the real mobile use cases.

This work addresses the above discussed problems and proposes a novel PyNET-V2 Mobile CNN architecture capable of processing 12MP photos directly on mobile devices while achieving high fidelity and visual results on the considered learned ISP task. The architecture is designed taking into account all major particularities of mobile AI accelerators such as limited computational power, memory restrictions and a constrained set of supported layers / ops, and is compatible with the latest NPUs and APUs. To evaluate the proposed solution, we use the Fujifilm UltraISP dataset containing real RAW images captured by mobile camera sensor and the target photos shot with a professional Fujifilm camera. As our main target is to replace the classical image signal processing pipelines, we perform a detailed comparison of the results obtained with the proposed solution and phone's built-in ISP system. Furthermore, we check the runtime of the PyNET-V2 Mobile model on real mobile GPUs and NPUs, and propose additional model variants for constrained low-power devices.


\begin{figure*}[th!]
\centering
\includegraphics[width=1\linewidth]{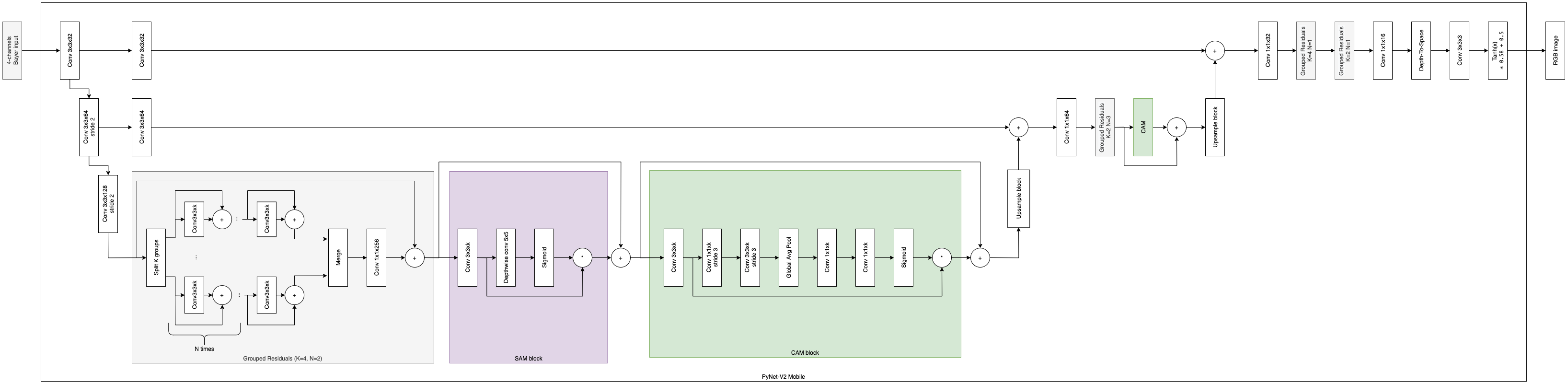}

\vspace{-0mm}
\caption{The overall architecture of the proposed PyNET-V2 Mobile model and its channel (CAM) and spatial (SAM) attention blocks. Best zoomed on screen.}
\label{fig:model_b_architecture}
\vspace{-2.0mm}
\end{figure*}

\section{Fujifilm UltraISP Dataset}
\label{sec:dataset}

In this paper, we use the Fujifilm UltraISP dataset collected using the Fujifilm GFX100 medium format 102 MP camera capturing the target high-quality images, and a popular Sony IMX586 Quad Bayer mobile camera sensor that can be found in tens of mid-range and high-end mobile devices released in the past 3 years. The Sony sensor was mounted on the \mbox{MediaTek} Dimensity 820 development board, and was capturing both raw and processed (by its built-in ISP system) 12MP images. The Dimensity board was rigidly attached to the Fujifilm camera, and they were shooting photos synchronously to ensure that the image content is identical. The dataset contains over 6 thousand daytime image pairs captured at a wide variety of places with different illumination and weather conditions. An example set of full-resolution photos from the Fujifilm UltraISP dataset is shown in Fig.~\ref{fig:dataset}.
As the collected RAW-RGB image pairs were not perfectly aligned, they were initially matched using the state-of-the-art deep learning based dense matching algorithm~\cite{truong2021learning} to extract 256$\times$256 pixel patches from the original photos. This procedure resulted in over 99K pairs of crops that were divided into training (93.8K), validation (2.2K) and test (3.1K) sets and used for model training and evaluation. It should be mentioned that all alignment operations were performed on Fujifilm RGB images only, therefore RAW photos from the Sony sensor remained unmodified, exhibiting the same values as read from the sensor.

\section{Architecture}
\label{sec:architecture}

When designing an architecture for the learned ISP task capable of processing high-resolution images on mobile devices and providing good visual results, one needs to address the following particularities of this problem:
\begin{itemize}
\item The designed model needs to perform both local (\eg, texture enhancement, noise reduction, super-resolution) and global (\eg, brightness, white balance and color rendition adjustments) image processing. The previously proposed solutions are either good at only one aspect (\eg, ResNet-~\cite{ledig2017photo}, U-Net-~\cite{ronneberger2015u}, DenseNet-\cite{huang2017densely} based models), or are too heavy for edge inference (\eg, PyNET~\cite{ignatov2020replacing}).
\item The network should be memory efficient to be able to process high-resolution images on mobile AI accelerators, where RAM size is usually very limited.
\item The model should only contain operators and layers supported by mobile AI accelerators such as NPUs or GPUs. In the latest Android R version, one is restricted to only 101 ops at maximum~\cite{NNAPIDrivers2021,NNAPI13Specs}, while on older system this number can be even lower than 28~\cite{NNAPI10Specs}.
\item Unlike large models such as PyNET~\cite{ignatov2020replacing} or Pix2Pix~\cite{isola2017image} with a size of hundreds of megabytes, the designed solution should be quite compact as it is usually bundled directly with the camera application.
\item The computational complexity of the model should be reasonably low in order to achieve an acceptable latency on mobile devices.
\end{itemize}
In this work, we address all the above limitations and propose the PyNET-V2 Mobile architecture the structure of which is inspired by the original PyNET~\cite{ignatov2020replacing} model, though the design of the network was fully revised in order to be compatible with mobile devices and AI accelerators. The overall model design is illustrated in Fig.~\ref{fig:model_b_architecture}, its description is provided below.

\begin{figure*}[t!]
\centering
\setlength{\tabcolsep}{1pt}
\resizebox{\linewidth}{!}
{
\begin{tabular}{ccccc}
\scriptsize{Visualized RAW Image}\normalsize  & \scriptsize{RAW Processed with Photoshop}\normalsize  & \scriptsize{MediaTek ISP Photo}\normalsize &\scriptsize{PyNet-V2 Mobile}\normalsize & \scriptsize{Fujifilm GFX 100 Photo}\normalsize\\
    \includegraphics[width=0.2\linewidth]{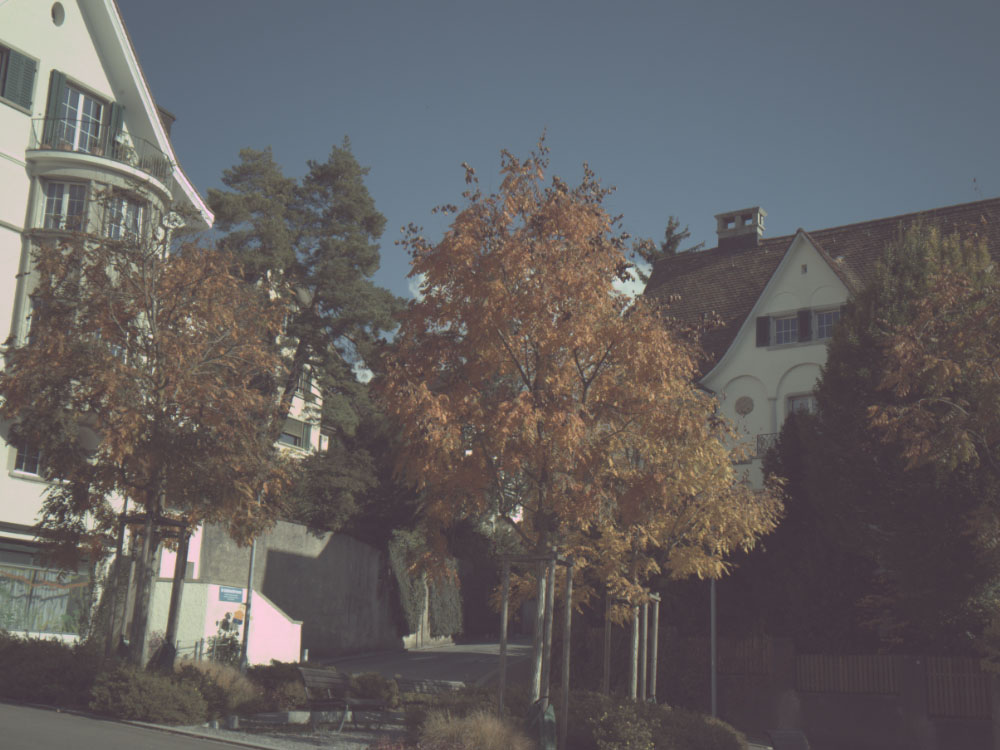}&
    \includegraphics[width=0.2\linewidth]{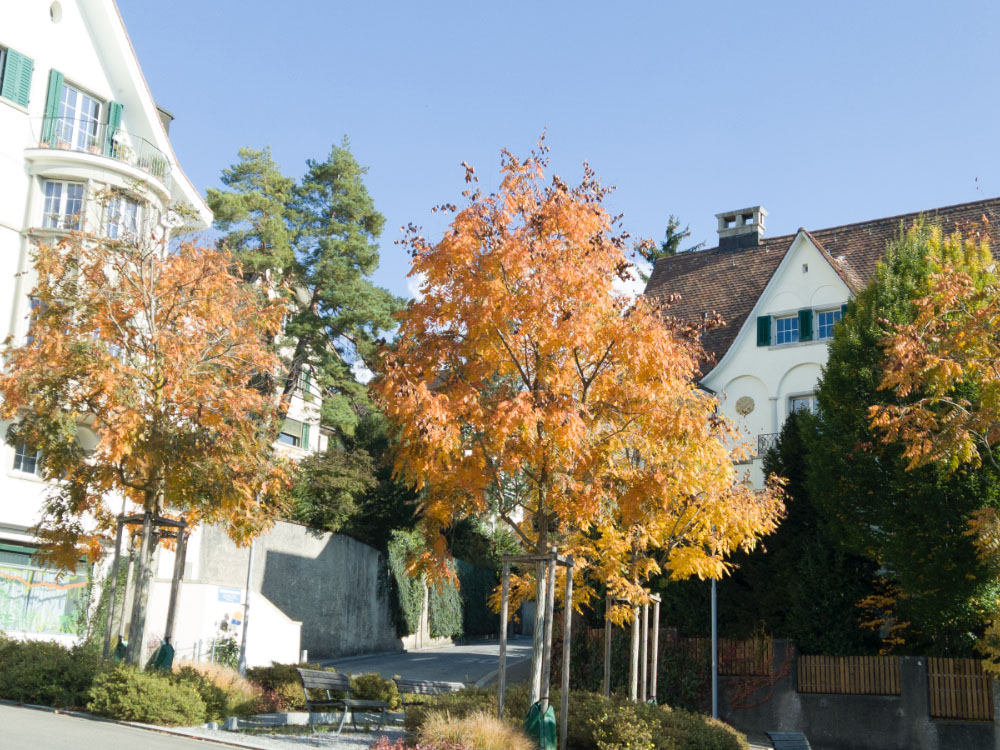}&
    \includegraphics[width=0.2\linewidth]{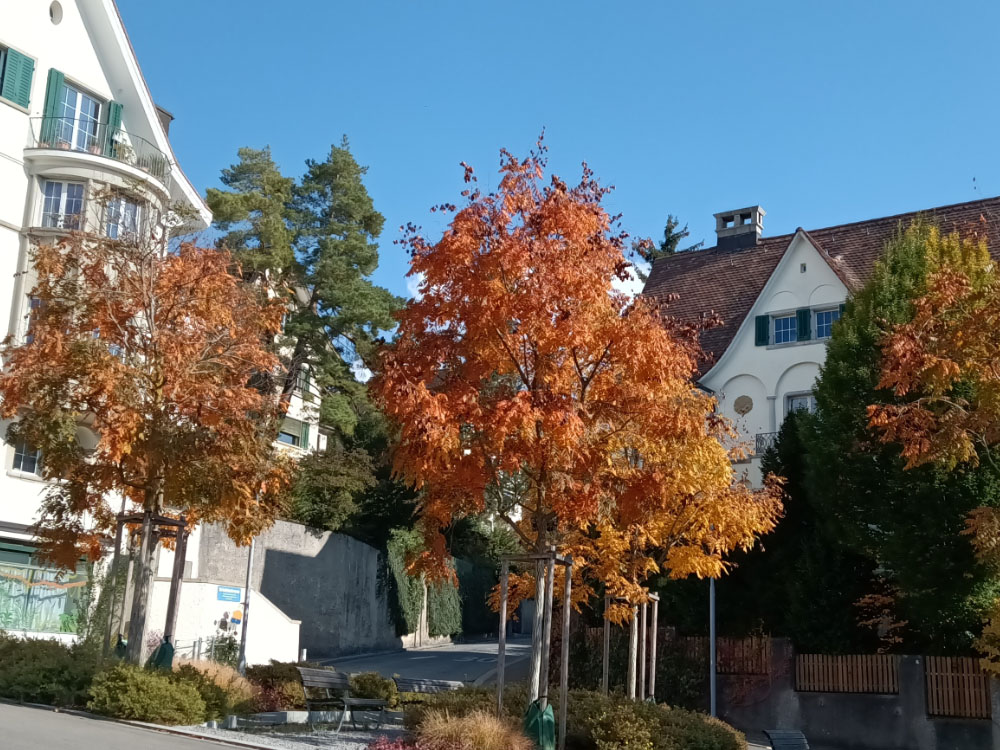}&
    \includegraphics[width=0.2\linewidth]{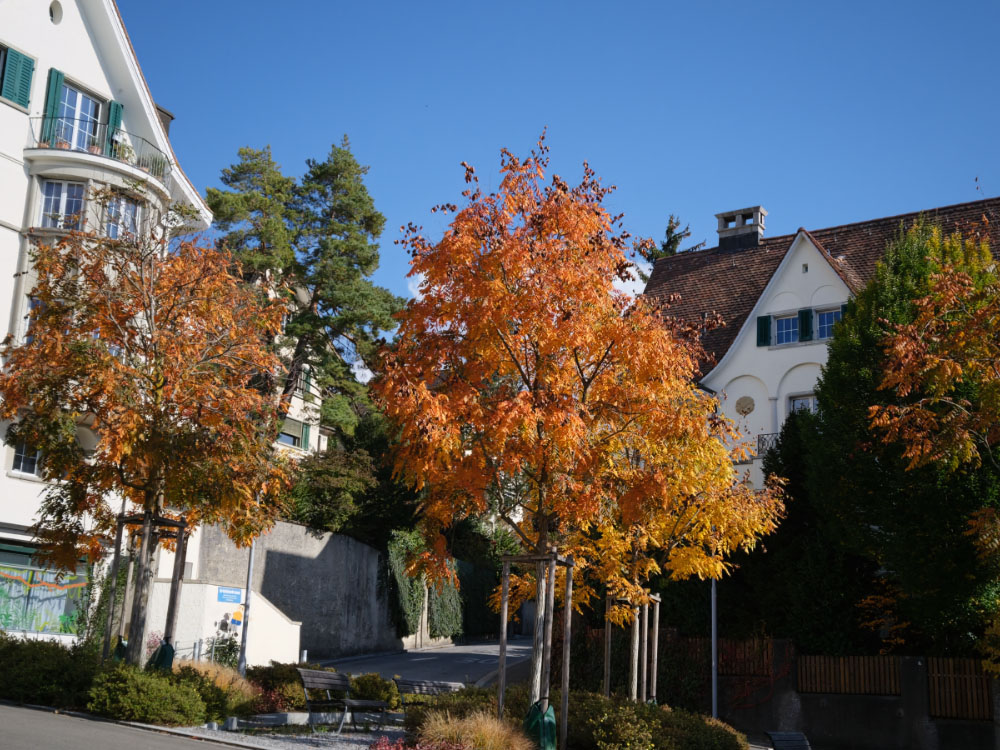}&
    \includegraphics[width=0.2\linewidth]{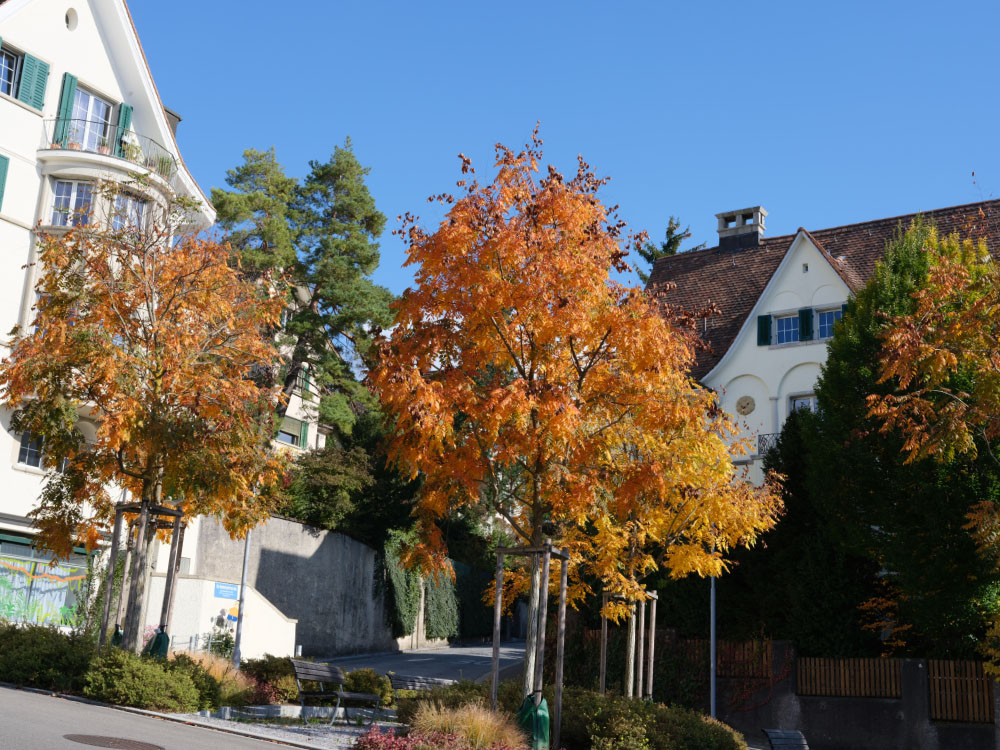} \\
    \includegraphics[width=0.2\linewidth]{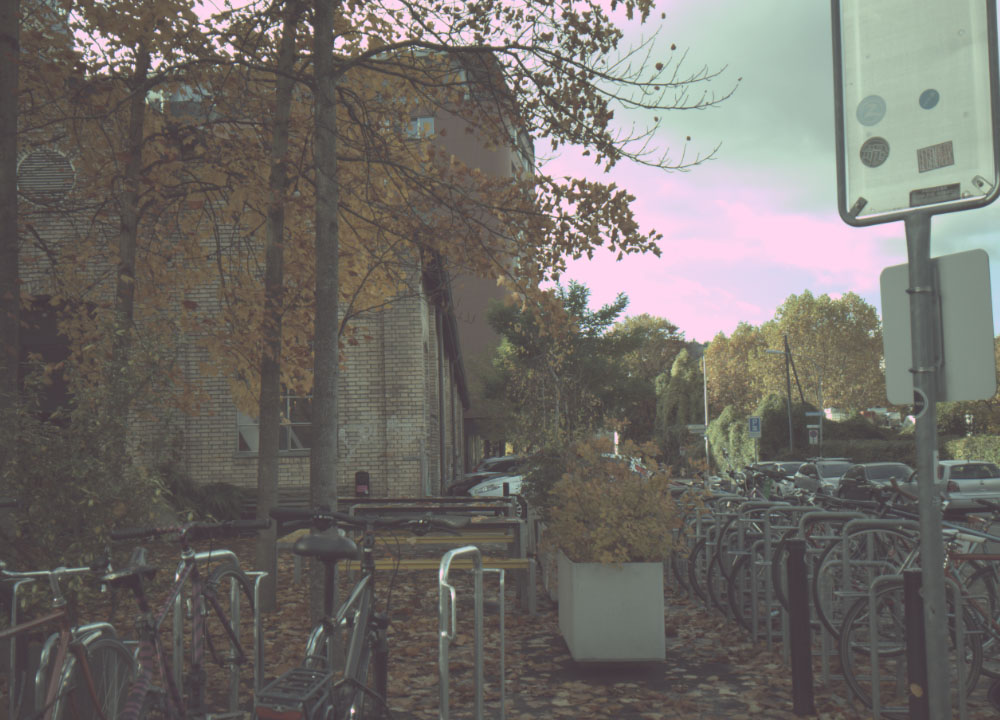}&
    \includegraphics[width=0.2\linewidth]{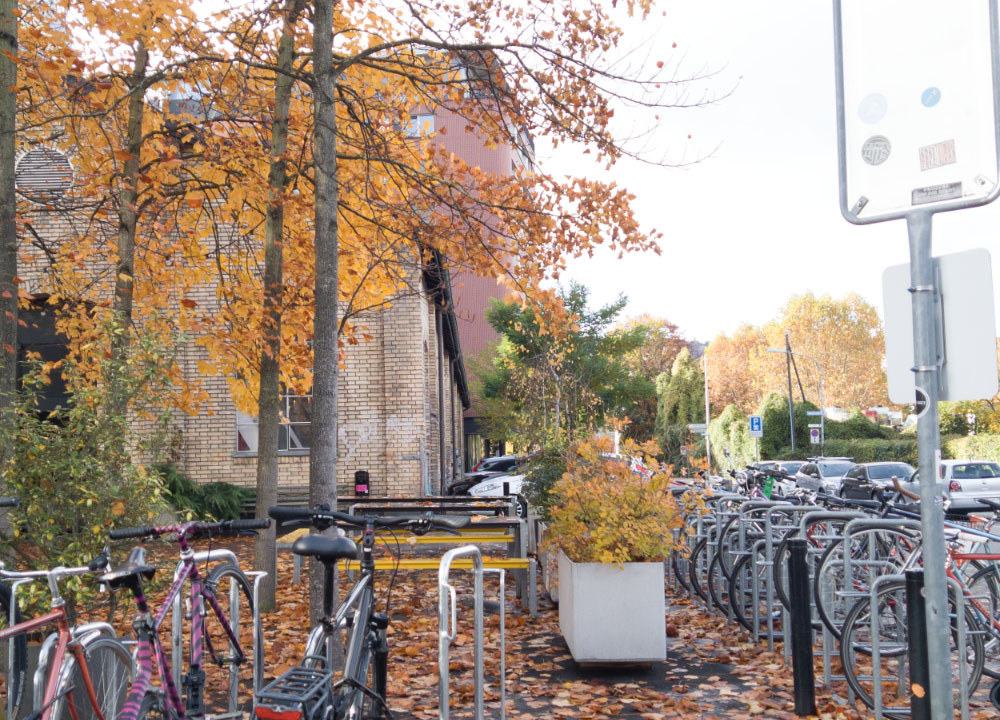}&
    \includegraphics[width=0.2\linewidth]{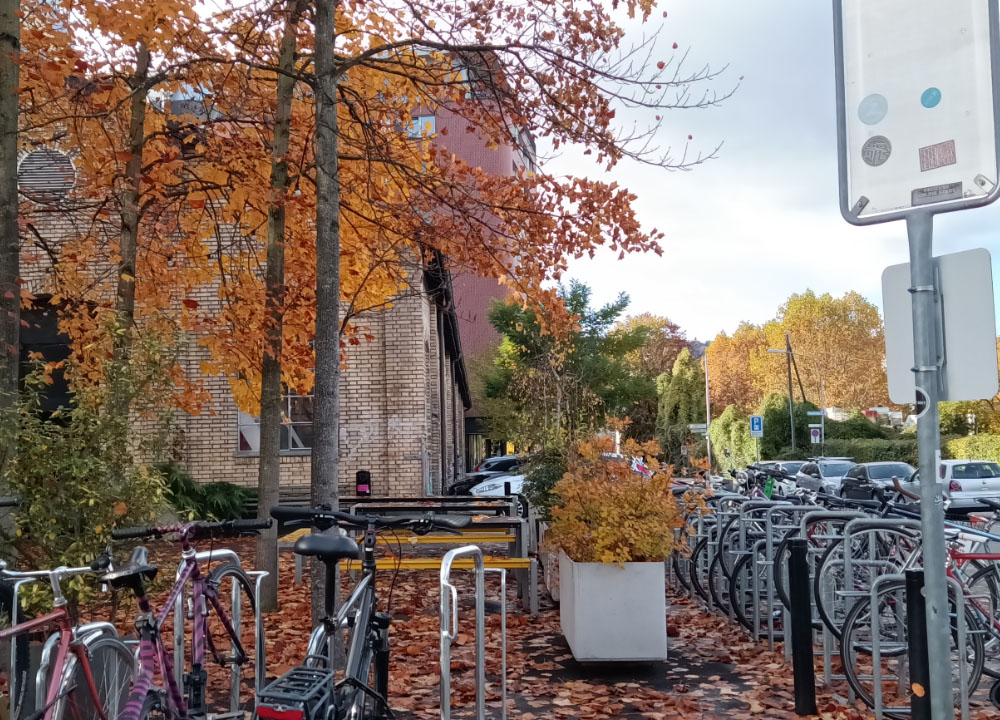}&
    \includegraphics[width=0.2\linewidth]{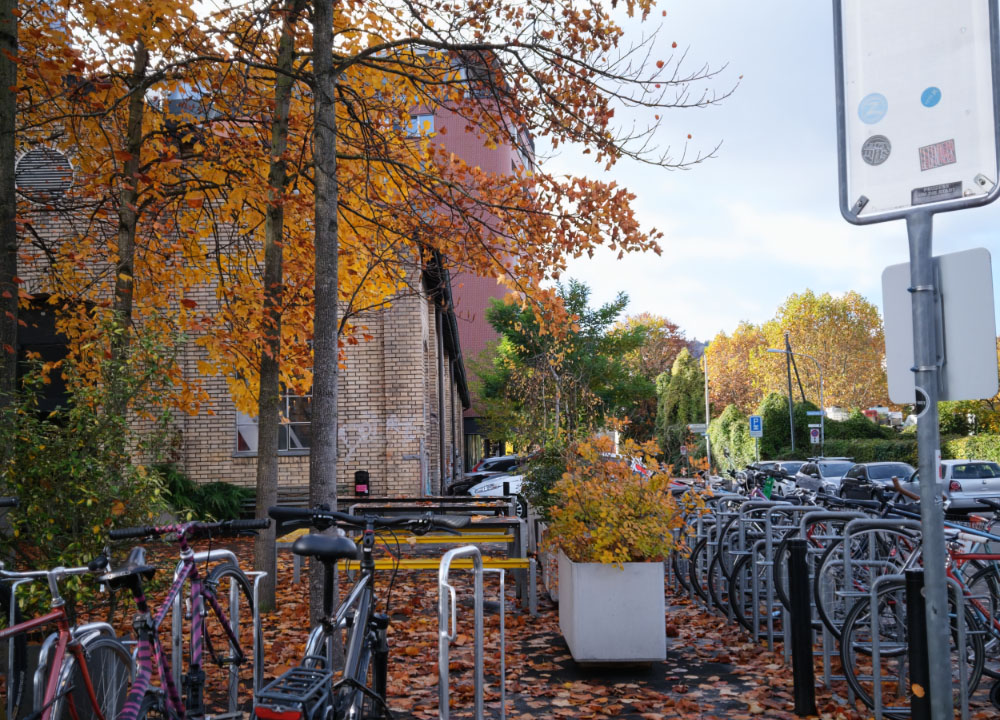}&
    \includegraphics[width=0.2\linewidth]{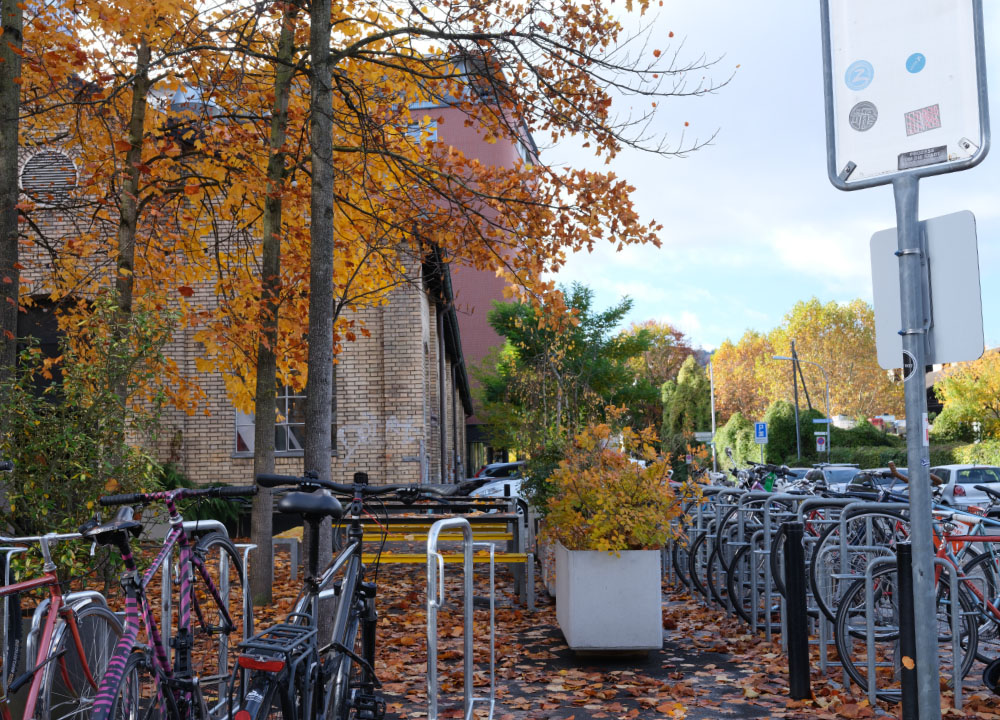}
\end{tabular}
}
\vspace{-0.2cm}
\caption{Sample visual results obtained with the proposed deep learning method. Best zoomed on screen.}
\vspace{-0.2cm}
\label{fig:example_photos_inference_1}
\end{figure*}

\begin{figure*}[t!]
\centering
\setlength{\tabcolsep}{1pt}
\resizebox{\linewidth}{!}
{
\begin{tabular}{cccccccccc}
    \includegraphics[width=0.1\linewidth]{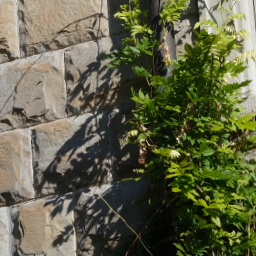}&
    \includegraphics[width=0.1\linewidth]{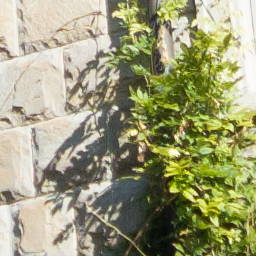}&
    \includegraphics[width=0.1\linewidth]{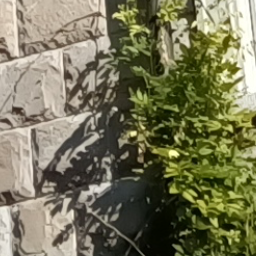}&
    \includegraphics[width=0.1\linewidth]{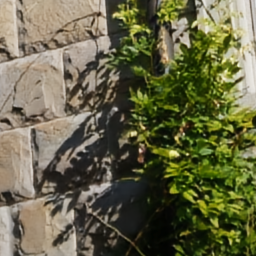}&
    \includegraphics[width=0.1\linewidth]{figs/2024_patch2_fuji.png} &
    \includegraphics[width=0.1\linewidth]{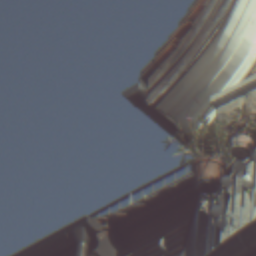}&
    \includegraphics[width=0.1\linewidth]{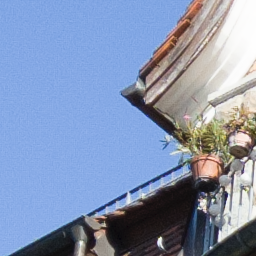}&
    \includegraphics[width=0.1\linewidth]{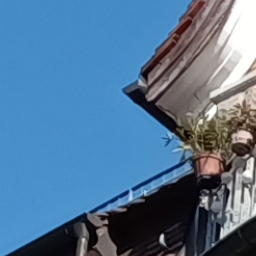}&
    \includegraphics[width=0.1\linewidth]{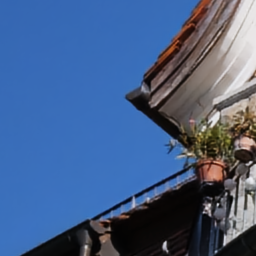}&
    \includegraphics[width=0.1\linewidth]{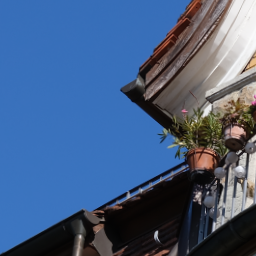} \\
    \includegraphics[width=0.1\linewidth]{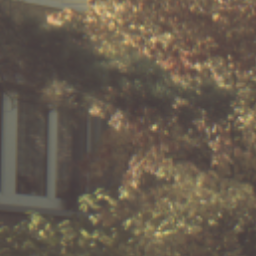}&
    \includegraphics[width=0.1\linewidth]{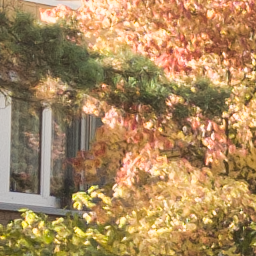}&
    \includegraphics[width=0.1\linewidth]{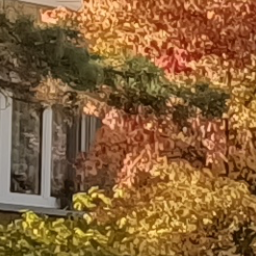}&
    \includegraphics[width=0.1\linewidth]{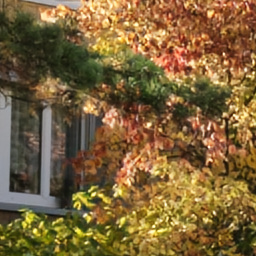}&
    \includegraphics[width=0.1\linewidth]{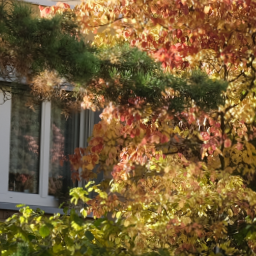} &
    \includegraphics[width=0.1\linewidth]{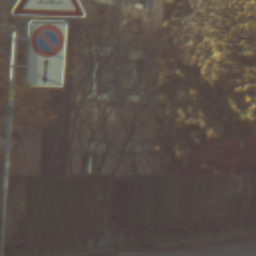}&
    \includegraphics[width=0.1\linewidth]{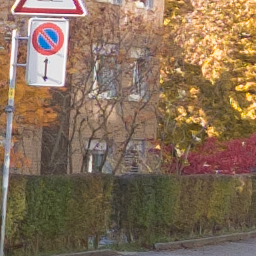}&
    \includegraphics[width=0.1\linewidth]{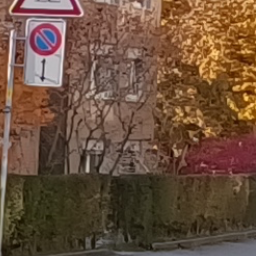}&
    \includegraphics[width=0.1\linewidth]{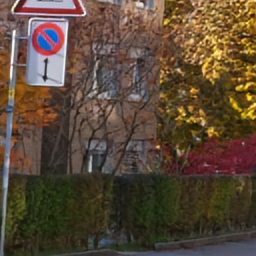}&
    \includegraphics[width=0.1\linewidth]{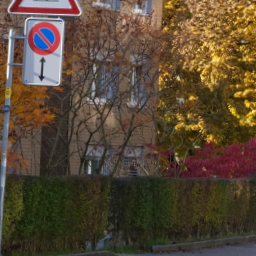}
\end{tabular}
}
\vspace{-0.2cm}
\caption{Sample image crops for several photo processing approaches. From left to right: visualized RAW image, RAW photo processed with Photoshop, image obtained with MediaTek's built-in ISP system, the result of the PyNet-V2 Mobile, and the target Fujifilm photo.}
\label{fig:example_photos_inference_2}
\vspace{-0.2cm}
\end{figure*}

\noindent \textBF{Overall image processing workflow.} The proposed model has an inverted pyramidal structure and consists of three different scales. The input data is processed sequentially at each scale, and the resulting features are then bilinearly upsampled and concatenated with the ones from a higher scale. The model accepts the raw RGBG Bayer data coming directly from the camera sensor. The input is then grouped in 4 feature maps corresponding to each of the four RGBG color channels using the space-to-depth op.

\noindent \textBF{Convolutional layers.} When designing a model for mobile devices, one should avoid using large convolutional filters due to the aforementioned memory limitations. Therefore, unlike in the original PyNET~\cite{ignatov2020replacing} paper, in the proposed architecture the number of convolutional filters is halved at each higher scale to reduce RAM consumption. Additionally, only $3\times 3$ convolutions are used at all scales (except for the attention blocks) as larger sizes lead to drastically increased memory usage. Each convolutional layer is followed by the \textit{PReLU} activation function except for the last one, where the \textit{Tanh} activation is used to map the outputs to the (-1, 1) interval.

\noindent \textBF{Grouped residual blocks.} To improve the model efficiency, we use residual blocks with grouped convolutions. The input feature maps in these blocks are split into several parts (from 2 to 4) and fed to separate convolutional channels working in parallel to decrease the computational costs. Additionally, instance normalization is applied to the outputs of each second channel. The obtained features are then concatenated and passed to the next layer.

\noindent \textBF{CAM block.} To ensure that the model has enough capacity to perform global image adjustments, we added several enhanced channel attention blocks with the structure shown in Fig.~\ref{fig:model_b_architecture}. In these blocks, the standard $3\times3$ convolutional layer is followed by a $1\times1$ one and one $3\times3$ convolution with stride 3 used to learn the global content-dependent features and reduce the resolution of the feature maps by 9 times. Finally, the average pooling op is used to get $1\times 1 \times \text{\textit{filter size}}$ features that are then passed to two additional conv layers generating channel normalization coefficients. This architecture is both performant and computationally efficient due to aggressive dimensionality reduction, leading to an execution time just slightly above that of a normal $3\times 3$ convolution.

\noindent \textBF{SAM block.} The model's performance was further boosted by using spatial attention modules. Each SAM block consists of one normal 3$\times$3 convolution and one depthwise convolution with a kernel size of 5$\times$5 followed by the sigmoid activation.

\noindent \textBF{Model operators.} The proposed PyNET-V2 Mobile model contains only layers supported by the Neural Networks API 1.2~\cite{NNAPI12Specs}, and thus can run on any NNAPI-compliant AI accelerator (such as NPU, APU, DSP or GPU) available on mobile devices with Android 10 and above.

\noindent \textBF{Model size and memory consumption.} The size of the PyNET-V2 Mobile network is 3.6 MB when exported for inference using the TFLite FP32 format. The model consumes around 0.3 and 1.4 GB of RAM when processing FullHD and 12MP photos on mobile GPUs, respectively.

\noindent \textBF{Training details.} To train the model, we follow the same sequential training scheme as in the PyNET papers~\cite{ignatov2020replacing,ignatov2020rendering}. This approach allows to achieve good semantically-driven reconstruction results at lower scales that are working with images of smaller resolution and thus performing mostly global image adjustments, while the higher scales are primarily learning to add the missing details and refine the quality of the reconstructed texture.

Each model layer was trained with a combination of the VGG-based~\cite{johnson2016perceptual} perceptual, SSIM and MSE losses in two steps. First, the network was optimized with a combination of the VGG-based and SSIM losses to learn the correct texture and edge reconstruction. Next, it was trained with a combination of the MSE and SSIM losses to enhance the quality of color rendering and brightness adjustments. The highest level was additionally trained with the 3rd step with the SSIM loss only to fine-tune the model performance.
A multi-iterative training process was used, meaning that the above steps were repeated one after the other several times till model convergence.

\noindent \textBF{Implementation details.} The model was implemented in TensorFlow and trained on one \textit{Nvidia GeForce RTX 2060} GPU with batch size 20. The network parameters were optimized using the ADAM~\cite{kingma2014adam} algorithm with a learning rate of $5e-5$ that was reduced on plateau till convergence. Random flips and rotations were applied to augment the training data and prevent model overfitting. After the end of the main training process, the network was additionally fine-tuned on training patches for which the PSNR score was 25dB and above, which improved the final results by 0.1 --- 0.15dB.

\section{Experiments}
\label{sec:experiments}

\begin{figure*}[t!]
\centering
\setlength{\tabcolsep}{1pt}
\resizebox{\linewidth}{!}
{
\begin{tabular}{ccccccccc}
\includegraphics[width=0.111\linewidth]{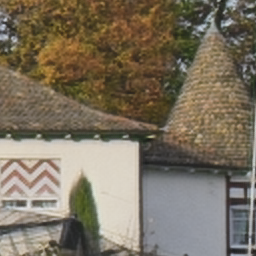}&
\includegraphics[width=0.111\linewidth]{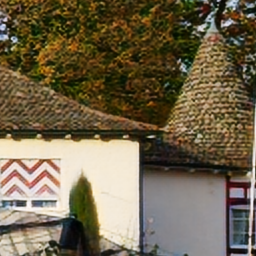}&
\includegraphics[width=0.111\linewidth]{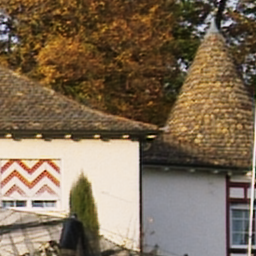}&
\includegraphics[width=0.111\linewidth]{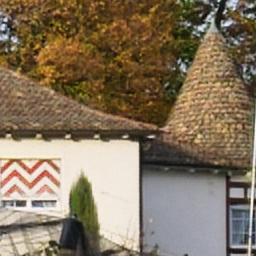}&
\includegraphics[width=0.111\linewidth]{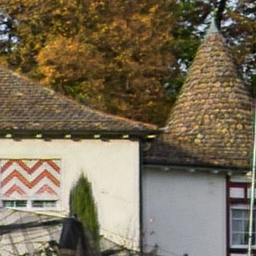}&
\includegraphics[width=0.111\linewidth]{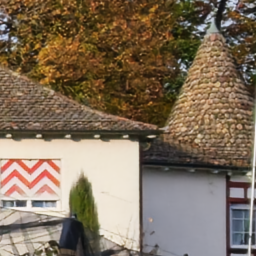}&
\includegraphics[width=0.111\linewidth]{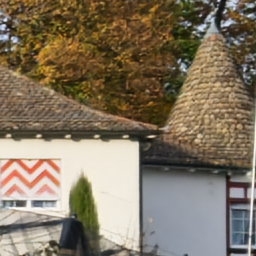}&
\includegraphics[width=0.111\linewidth]{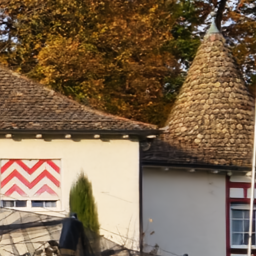}&
\includegraphics[width=0.111\linewidth]{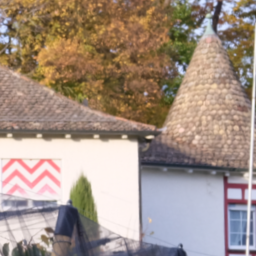} \\
\includegraphics[width=0.111\linewidth]{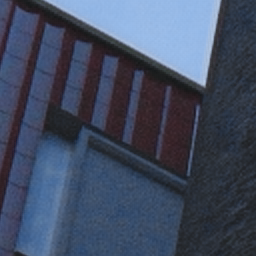}&
\includegraphics[width=0.111\linewidth]{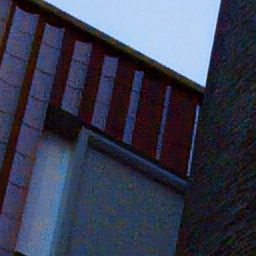}&
\includegraphics[width=0.111\linewidth]{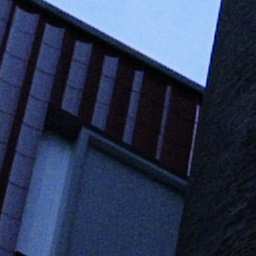}&
\includegraphics[width=0.111\linewidth]{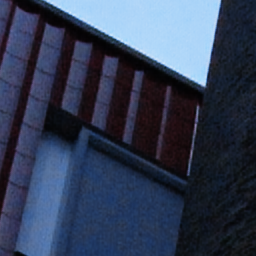}&
\includegraphics[width=0.111\linewidth]{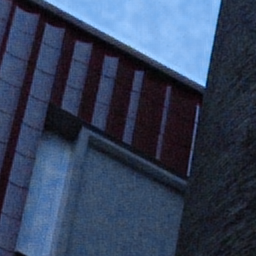}&
\includegraphics[width=0.111\linewidth]{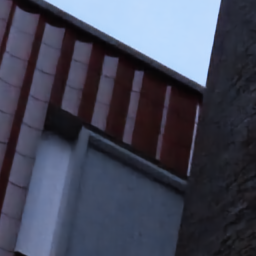}&
\includegraphics[width=0.111\linewidth]{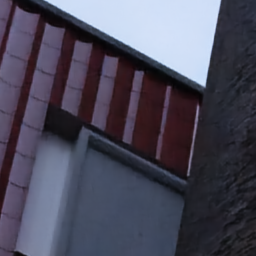}&
\includegraphics[width=0.111\linewidth]{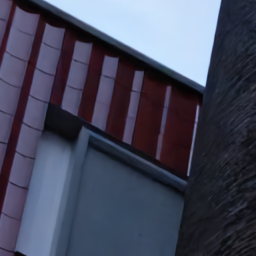}&
\includegraphics[width=0.111\linewidth]{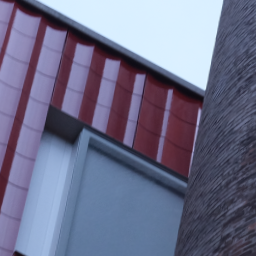} \\
\includegraphics[width=0.111\linewidth]{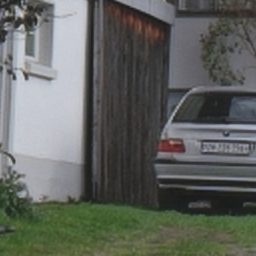}&
\includegraphics[width=0.111\linewidth]{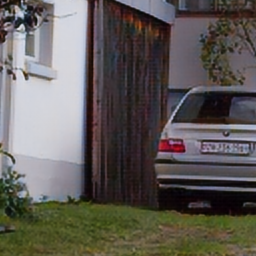}&
\includegraphics[width=0.111\linewidth]{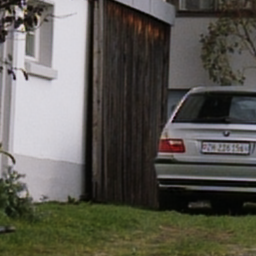}&
\includegraphics[width=0.111\linewidth]{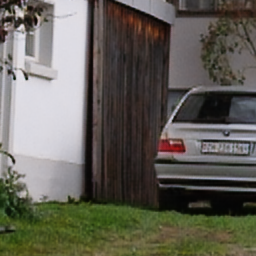}&
\includegraphics[width=0.111\linewidth]{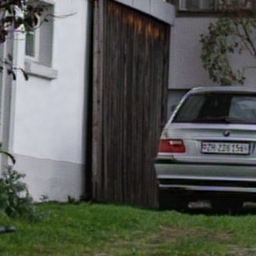}&
\includegraphics[width=0.111\linewidth]{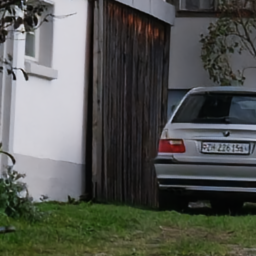}&
\includegraphics[width=0.111\linewidth]{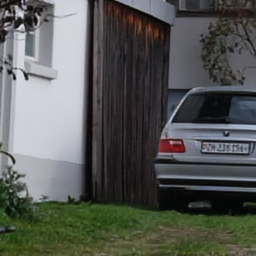}&
\includegraphics[width=0.111\linewidth]{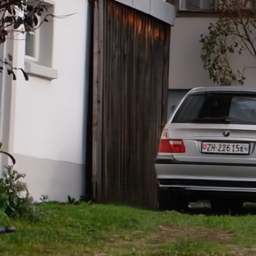}&
\includegraphics[width=0.111\linewidth]{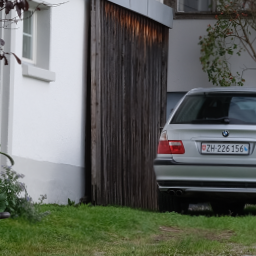} \\
\end{tabular}
}
\vspace{-0.2cm}
\caption{Comparison of the visual results obtained with different deep learning based ISP solutions. From left to right: FSRCNN~\cite{dong2016accelerating}, FPIE~\cite{de2018fast}, ENERZAi~\cite{ignatov2021learned}, CSANet~\cite{hsyu2021CSANet}, SmallNet~\cite{ignatov2021learned}, DPED~\cite{ignatov2017dslr}, our PyNET-V2 Mobile architecture, original PyNet~\cite{ignatov2020replacing} model, and the target Fujifilm photo.}
\label{fig:solutions_comparison}
\vspace{-2.0mm}
\end{figure*}

In this section, we evaluate the proposed PyNET-V2 Mobile architecture on the real Fujifilm UltraISP dataset and mobile devices to answer the following questions:
\begin{itemize}
\itemsep -0.0em
\item How good is the quality of the RGB images reconstructed with the PyNET-V2 Mobile compared to the classical hand-crafted ISP approaches;
\item How well this solution performs compared to the state-of-the-art deep learning solutions tuned for this task;
\item What is the runtime of this model when performing the inference on mobile GPUs and AI accelerators;
\item What are the limitations of the proposed solution.
\end{itemize}
To answer these questions, we performed a wide range of experiments which results are described in the next sections.

\subsection{Qualitative evaluation}

We start the performance analysis of the proposed solution with an inspection of the visual results obtained using this deep learning-based approach. Figure~\ref{fig:example_photos_inference_1} shows sample RGB images reconstructed with the PyNET-V2 Mobile model together with the original visualized RAW photos, images processed with Photoshop's RAW photo processing engine, images obtained with MediaTek's built-in ISP system, and the target photos from the Fujifilm camera. The first observation is that the proposed neural network performed an accurate reconstruction of the RGB image, with precise color rendition and white balancing. No notable issues can be observed at both global and local levels, the recovered photo does not exhibit any textural or color artifacts even when rendering complex image areas. The dynamic range of the reconstructed photos is high and close to the one of the target Fujifilm images.

\begin{table}[t!]
\centering
\resizebox{1.0\linewidth}{!}
{
\begin{tabular}{l|cccc|cc}
\hline
Method \, & \, PSNR \, & \, SSIM \, & \, FOM \, & \, LPIPS \, &\multicolumn{2}{c}{MediaTek Dimensity 1000+ GPU} \\
 \, & \,  \, & \,  &  &  & \, FullHD, ms \, & \, 12MP, ms \, \\
\hline
\hline
SmallNet~\cite{ignatov2021learned}  & 23.2 & 0.8467 & 0.7305 & 0.2870 & \textBF{18.4} & \textBF{100} \\
FPIE~\cite{de2018fast} & 23.23 & 0.8481 & 0.7312 & 0.2505 & 208 & 1138  \\
FSRCNN~\cite{dong2016accelerating} & 23.27 & 0.8303 & 0.6792 & 0.3027 & 40.8  & 232 \\
ENERZAi~\cite{ignatov2021learned} & 23.41 & 0.8534 & 0.6932 & 0.2867 & 31.2  & 123 \\
CSANet~\cite{hsyu2021CSANet}  & 23.73 & 0.8487 & 0.7174 & 0.2552 & 44.2  & 241 \\
\hline
DPED~\cite{ignatov2017dslr}  & 24.56 & 0.8872 & 0.7445 & 0.2111 & 658 & 4027 \\
PyNet~\cite{ignatov2020replacing} & \textBF{25.01} & \textBF{0.8985} & \textBF{0.7528} & \textBF{0.1871} & 12932 & \small{OOM} \\
\hline
\textit{PyNET-V2 Mobile} & 24.72 & 0.8783 & 0.7283 & 0.2164 & 274 & 1492 \\
\end{tabular}
}
\vspace{0.0mm}
\caption{Average PSNR / SSIM scores on test crops and the runtime results on the MediaTek Dimensity 1000+ mobile platform for several deep learning based ISP solutions. OOM stands for the ``out-of-memory'' exception thrown during the inference.}
\label{tab:comparison_to_baselines}
\vspace{-7.0mm}
\end{table}

A more detailed comparison to the results obtained with MediaTek's built-in ISP system revealed that the proposed neural network-based approach can yield more natural looking photos at local texture scale (Fig.~\ref{fig:example_photos_inference_2}). In particular, the images processed with the considered ISP system exhibit a strong ``watercolor'' effect caused by combining aggressive image sharpening and edge enhancement filters with heavy image denoising. In contrast, the photos obtained with the PyNET-V2 Mobile do not show any signs of oversharpening or artificially altered texture. In terms of the true image resolution, the proposed solution demonstrates higher results, being able to reconstruct small details more precisely. The same also applies to color rendition, which is much closer to the Fujifilm images.

When comparing the results against the images processed with Photoshop, one might notice that the latter ones preserve more local details at the expense of a considerably larger amount of noise even in bright image areas. When applying Photoshop's built-in noise suppression algorithms, the difference becomes negligible. It should be also noted that, as expected, the target Fujifilm photos significantly outperform the results of the ISP pipeline, Photoshop and the proposed solution in all aspects, especially in terms of resolution.

\subsection{Quantitative evaluation}

\begin{table*}[t!]
\centering
\resizebox{\linewidth}{!}
{
\begin{tabular}{l|cc|cc|cc|c}
\hline
Mobile SoC & \,  Dimensity 9000 \,  & \,  Dimensity 820 \,  & \,  Exynos 2100 \,  & \, Exynos 990 \, & \, Kirin 9000 \, & \, Snapdragon 888 \, & \, Google Tensor \\
GPU & \, \small Mali-G710 MC10, ms \, & \, \small Mali-G57 MC5, ms \, & \, \small Mali-G78 MP14, ms \, & \, \small Mali-G77 MP11, ms \, & \, \small Mali-G78 MP24, ms \, & \, \, \small Adreno 660, ms \, & \, Mali-G78 MP20, ms \\
\hline
Full HD & 224  & 430  & 206  & 265  & 192 & 275  & 197  \\
12MP    & 1246 & 2327 & 1056 & 1382 & 963 & 1542 & 1091 \\
\end{tabular}
}
\vspace{2.0mm}
\caption{The speed of the proposed PyNET-V2 Mobile architecture on several popular mobile GPUs for different photo resolutions.
The runtime was measured with the AI Benchmark application using the TFLite GPU delegate~\cite{lee2019device}.
}
\label{tab:runtime_on_socs}
\vspace{-2.0mm}
\end{table*}

Since the proposed network architecture was developed for performant and computationally efficient on-device photo processing, in this section we compare its numerical, visual and runtime results against the previously introduced deep learning-based solution for photo processing and enhancement. The following models are used in the next experiments:
\begin{itemize}
\item PyNET~\cite{ignatov2020replacing}: a recent state-of-the-art solution for end-to-end learned ISP problem.
\item DPED~\cite{ignatov2017dslr}: a ResNet-based architecture for mobile photo enhancement.
\item FPIE~\cite{de2018fast}: an enhanced DPED~\cite{ignatov2017dslr}-based neural network optimized for fast on-device image processing.
\item FSRCNN~\cite{dong2016accelerating}: a popular computationally efficient model used for various image enhancement problems.
\item Compressed U-Net~\cite{ignatov2021learned}: a U-Net~\cite{ronneberger2015u} based model with hardware-specific adaptations for edge inference.
\item ENERZAi~\cite{ignatov2021learned}~--- a model designed for efficient image ISP, derived from the ESRGAN~\cite{wang2018esrgan} architecture.
\item CSANet~\cite{hsyu2021CSANet}: an NPU-friendly architecture developed for the learned smartphone ISP problem.
\item SmallNet~\cite{ignatov2021learned}~--- a fast FSRCNN~\cite{dong2016accelerating} based architecture optimized for the learned smartphone ISP task.
\end{itemize}
All models were trained on the Fujifilm UltraISP dataset, their PSNR, SSIM, FOM~\cite{pinho1995edge} and LPIPS~\cite{zhang2018unreasonable} scores on the test image subset are reported in Table~\ref{tab:comparison_to_baselines}, sample visual results for all methods are illustrated in Fig.~\ref{fig:solutions_comparison}. As we are targeting on-device RAW photo processing, we additionally measured the runtime of all solutions on FullHD and 12MP images on the MediaTek Dimensity 1000+ mobile SoC. For this, we used the publicly available AI Benchmark application~\cite{ignatov2018ai,ignatov2019ai} that allows to load any custom TensorFlow Lite model and run it on any Android device with various acceleration options including GPU, NPU, DSP and CPU inference. The models were accelerated on the Mali-G77 GPU of the above mentioned Dimensity SoC as this option delivered the best latency for all architectures. The resulted runtime values are reported in Table~\ref{tab:comparison_to_baselines}.

The results demonstrate that the proposed PyNET-V2 Mobile is able to achieve good image reconstruction results while being able to process one FullHD and 12MP on the target platform under 0.3 and 1.5 seconds, respectively. In terms of PSNR scores, it outperformed all other solutions except for the original PyNET architecture. As expected, the latter one delivers higher numerical results, though its computational complexity is infeasible for processing photos on mobile devices: it demonstrates an almost 50 times larger inference time when processing FullHD resolution images on the Dimensity 1000+ chipset, while for 12MP photos the inference fails with the \textit{out-of-memory} exception. Thus, its results were provided mainly for the reference as it cannot be practically used for on-device photo processing.

When compared to the DPED model, one can notice that the latter produces sharper images which is also reflected by its higher SSIM and FOM scores. However, the photos reconstructed with this architecture are not always ideal in terms of brightness and tone mapping as it is processing the images at the original scale only, which makes it hard to perform global image adjustments. This also leads to its lower PSNR score on the considered dataset. The PyNET-V2 Mobile model is additionally more computationally efficient than the DPED architecture, being able to process photos up to 2.5 times faster on the same hardware.

As for the rest of the solutions, they significantly fall behind both numerically and visually compared to the PyNET-V2 Mobile, PyNET and DPED models despite showing good latency. In particular, none of the considered models was able to suppress noise present on the original photos or perform an accurate color reconstruction. More importantly, almost all of them show severe checkerboard artifacts and amplified color noise. Therefore, though they still might be useful for performing some preliminary rapid photos processing (\eg, for demonstrating photo thumbnail), they are not suitable for generating the final high-quality image reconstruction results. Thus, we can conclude that the proposed architecture closes the gap between the fast models providing low-quality visual results and the state-of-the-art solutions demanding computational resources that exceed the capacity of mobile devices, being perceptually much closer to the later ones.

\subsection{Runtime evaluation on mobile GPUs}

To explore the feasibility of high-resolution photo processing with the PyNET-V2 Mobile architecture, we additionally checked its runtime on all popular high-end mobile chipsets, and report the obtained results in Table~\ref{tab:runtime_on_socs}. As one can see, on the recent flagship SoCs like the Google Tensor, Exynos 2100 and Kirin 9000, it takes approximately one second to reconstruct one 12MP photo. On the mid-range MediaTek Dimensity 820 chipset, the latency increases to 2.4 seconds, which still fits in the runtime limits used in commercial devices. The results on smaller FullHD photos show that it can also be used for fast photo preview generation, requiring less 500 ms for image processing on all mobile platforms.

\subsection{Inference on Mobile NPUs}

\begin{figure*}[t!]
\centering
\setlength{\tabcolsep}{1pt}
\resizebox{1.0\linewidth}{!}
{
\begin{tabular}{cccc}
    \includegraphics[width=0.125\linewidth]{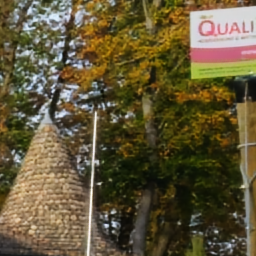}&
    \includegraphics[width=0.125\linewidth]{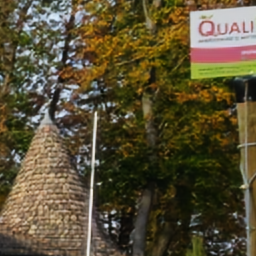}&
    \includegraphics[width=0.125\linewidth]{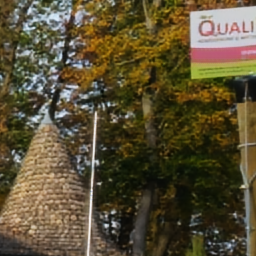}
    \includegraphics[width=0.125\linewidth]{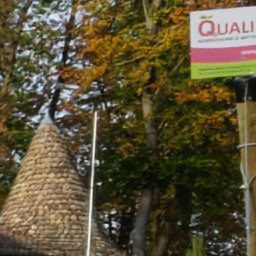}&
\end{tabular}
}
\vspace{-0.2cm}
\caption{Sample image reconstruction results for several PyNET-V2 Mobile variants: without instance normalization, with additional 2$\times$ input downsampling, with input downsampling and a modified upsampling block, and the original implementation.}
\label{fig:example_variants}
\vspace{-0.2cm}
\end{figure*}

While the above results demonstrate the feasibility of high-resolution photo processing with the PyNET-V2 Mobile architecture on smartphone GPUs, its latency can be substantially improved by running it on dedicated AI accelerators found in the majority of recent mobile devices. In this section, we evaluate the performance of this model on MediaTek's latest mobile platform, \emph{Dimensity 9000}, which features a powerful AI Processing Unit (APU) designed specifically for complex computer vision and image processing tasks. Table~\ref{tab:apu_performance} shows the runtime and power consumption results obtained on this chipset for 12MP images when running the PyNET-V2 Mobile model on GPU and APU. The considered AI accelerator was able to execute the entire floating-point model without any partitioning, and demonstrated an almost 2 times lower runtime and 3 times higher power efficiency for the same floating-point network compared to GPU inference. Moreover, these numbers can be further improved by performing model quantization and converting it to INT16 or INT8 formats. In the latter case, it requires slightly more than 400 ms for processing 12MP photo and offers an even larger reduction of power consumption, which is especially crucial when shooting burst photos as prevents fast battery draining.

\begin{table}[t]
\centering
\resizebox{1.0\linewidth}{!}
{
\begin{tabular}{l|cccc}
\hline
 \, & \multicolumn{4}{c}{MediaTek Dimensity 9000 mobile chipset} \\
& GPU (FP16) & APU (FP16) & APU (INT16) & APU (INT8) \\
\hline
\hline
Runtime, ms & 1246 & 767 & 645 & \textBF{413} \\
Power, fps/watt\, & 0.10 & 0.33 & 0.30 & \textBF{0.46} \\
\end{tabular}
}
\vspace{2.0mm}
\caption{The runtime and power consumption of the proposed model on the MediaTek Dimensity 9000 mobile platform.}
\label{tab:apu_performance}
\vspace{-4.0mm}
\end{table}

\subsection{Further model optimizations}

\begin{table}[t!]
\centering
\resizebox{0.96\columnwidth}{!}
{
\begin{tabular}{l|c|c|c}
Method & PSNR & SSIM & Runtime, ms  \\
\hline
\hline
PyNET-V2 Mobile NoNorm & 24.52 & 0.8705 & 1227 \\
PyNET-V2 Mobile Slim & 24.41& 0.8729  & 961 \\
PyNET-V2 Mobile Slim+ & 24.30 & 0.8691 & 808 \\
\hline
PyNET-V2 Mobile & 24.72 & 0.8783 & 1492 \\
\end{tabular}
}
\vspace{3.2mm}
\caption{Average PSNR / SSIM scores on test images and the runtime results on the Dimensity 1000+ GPU for several PyNET-V2 Mobile variants: without instance normalization, with additional 2$\times$ input downsampling, with input downsampling and a modified upsampling block, and the original implementation.
\label{tab:model_variants}}
\vspace{-2.2mm}
\end{table}

Though the proposed architecture shows good latency on mobile devices, one might want to further decrease its computational complexity in order to get a better runtime or be able to deploy it on low-power hardware. For this, we propose three different PyNET-V2 Mobile design modifications targeted at lowering the computational constraints:

\smallskip

\noindent \textBF{PyNET-V2 Mobile NoNorm:} a model without instance normalization layers. While this change has only a little effect on the runtime when executing the model on GPU, not all NPUs / APUs can run the instance norm op efficiently, especially when processing high-resolution data. Thus, disabling this op might potentially lead to significant inference speed increase on legacy or constrained AI accelerators.

\smallskip

\noindent \textBF{PyNET-V2 Mobile Slim:} a model performing all computations at twice lower resolution. This is achieved by using a strided convolution with stride 2 in the first layer, and performing an additional 2 times pixel shuffle at the end of the model. To avoid the information loss caused by input downsampling, the number of feature maps in conv layers is doubled in this network.

\smallskip

\noindent \textBF{PyNET-V2 Mobile Slim+:} A modification of the previous architecture that is using an optimized upscaling block: instead of the standard conv layer, depthwise convolution is applied after the bilinear upsamling operator.

\smallskip

Table~\ref{tab:model_variants} demonstrates the fidelity and runtime results of these architecture variants on the Dimensity 1000+ GPU, while Figure~\ref{fig:example_variants} shows sample image reconstruction results for the considered models. In terms of visual results, the images reconstructed with these networks are looking very similar, the difference mainly comes from slightly degraded image sharpness in the models performing 2 times image downsampling, a bit larger noise levels on low-light image areas, and less precise brightness prediction. One can also notice that the texture quality obtained with the PyNET-V2 Mobile Slim+ architecture degrades a bit, though this difference can be observed only when zooming in the photos significantly. As for the speed of these variants, the PyNET-V2 Mobile Slim version demonstrates a 35\% lower latency when running on GPU, while the Slim+ modification reduces the runtime by more than 1.8 times, requiring around 800ms to process one 12MP photos on the Dimensity SoC. The latency of the model without instance normalization also improves a bit, though, as expected, in this case the change is only around 15\% as the inference is performed on GPU that can execute this op efficiently.

\subsection{Limitations}

After analyzing the visual results on several hundred full-resolution images, we do not observe any global reconstruction issues, though on appr. 2-3\% of photos the white balance or brightness adjustments might not be very precise, leading to darker images or photos with pinkish / yellowish tints. Additionally, as the model was trained on day-time images, the reconstruction results on night photos are not perfect as the corresponding RAW data contains a large amount of noise which is not completely suppressed. Another limitation is that the network is not fixing the vignetting caused by camera optics, though this is a standard problem that can be completely resolved by applying conventional correction algorithms. Finally, as was noted in the previous sections, though the resolution of the reconstructed photos is higher than that of the corresponding ISP images, one could still further improve it and even to learn an additional 2 times photo super-resolution, since the target Fujifilm images allow for this, although a heavier model with a larger number of filters might be needed to solve this task efficiently.

\subsection{MicroISP Model}

Besides the PyNET-V2 Mobile model, we also developed a considerably faster and efficient MicroISP architecture that can process up to 32MP photos on mobile AI accelerators. The MicroISP network achieves a PSNR score of 23.87 dB on the considered FujiFilm UltraISP dataset, and its runtime on the Dimensity 9000 SoC is less than 650 ms when processing raw 32MP resolution images. A detailed description of this architecture and its results can be found in paper~\cite{ignatov2022microisp}.

\section{Conclusions}
\label{sec:conclusion}

We proposed the novel PyNET-V2 Mobile architecture for the learned smartphone ISP problem, yielding both good visual reconstruction results and low latency on mobile devices. Our solution learns to perform this task directly from the data in an end-to-end fashion, not requiring any manual supervision or hand-designed features. To check the performance of the model, we conducted experiments on the real Fujifilm UltraISP dataset that demonstrated its advantage over the traditional ISP systems as well as professional photo editing software. Compared to the previously proposed deep learning-base networks, the PyNET-V2 Mobile model is able to achieve both high visual quality and low latency, requiring only 1 second to process 12MP images on the recent mobile GPUs. Furthermore, the architecture is compatible with dedicated smartphone AI accelerators, enabling a further reduction in latency when inferring on NPUs or APUs. Finally, three additional model modifications further lower the computational complexity and create compatibility with legacy and low-power mobile hardware.

{\small
\bibliographystyle{IEEEtran}
\bibliography{egbib}
}

\end{document}